\documentclass{article}
\usepackage{arxiv}
\usepackage[utf8]{inputenc} 
\usepackage[T1]{fontenc}    
\usepackage{hyperref}       
\usepackage{url}            
\usepackage{booktabs}       
\usepackage{amsfonts}       
\usepackage{nicefrac}       
\usepackage{microtype}      
\usepackage{lipsum}		
\usepackage{graphicx}
\usepackage[square,sort,comma,numbers]{natbib}
\usepackage{doi}
\usepackage{amsmath}
\usepackage{xcolor}

\title{JoyVASA: Portrait and Animal Image Animation with Diffusion-Based Audio-Driven Facial Dynamics and Head Motion Generation
}
\date{}
\author{
    Xuyang Cao \thanks{Equal Contribution}\\
    JD Health International Inc. \\
    \texttt{newxuyangcao@gmail.com} \\
    \And
    Guoxin Wang \textsuperscript{*} \\
    Zhejiang University  \\
    JD Health International Inc. \\
    \texttt{guoxin.wang@zju.edu.cn}
    \And
    Sheng Shi \textsuperscript{*} \\
    JD Health International Inc.\\
    \texttt{shengshi\_neu@icloud.com} \\
    \And
    Jun Zhao  \\
    JD Health International Inc.\\
    \texttt{zhaojun10@jd.com} \\
    \And
    Yang Yao \\
    JD Health International Inc. \\
    \texttt{y-yao@outlook.com} \\
    \And
    Jintao Fei \\
    JD Health International Inc. \\
    \texttt{fei-jintao@outlook.com} \\
    \And
    Minyu Gao \\
    JD Health International Inc. \\
    \texttt{minyu.gao@outlook.com} \\
    \And
    Pei Xie \\
    JD Health International Inc. \\
    \texttt{xiepei13@jd.com} \\
}

\renewcommand{\shorttitle}{JoyVASA}

\begin{document}
\maketitle

\begin{abstract}
Audio-driven portrait animation has made significant advances with diffusion-based models, improving video quality and lip-sync accuracy. However, the increasing complexity of these models has led to inefficiencies in training and inference, as well as constraints on video length and inter-frame continuity. In this paper, we propose JoyVASA, a diffusion-based method for generating facial dynamics and head motion in audio-driven facial animation. Specifically, in the first stage, we introduce a decoupled facial representation framework that separates dynamic facial expressions from static 3D facial representations. This decoupling allows the system to generate longer videos by combining any static 3D facial representation with dynamic motion sequences. Then, in the second stage, a diffusion transformer is trained to generate motion sequences directly from audio cues, independent of character identity. Finally, a generator trained in the first stage uses the 3D facial representation and the generated motion sequences as inputs to render high-quality animations. With the decoupled facial representation and the identity-independent motion generation process, JoyVASA extends beyond human portraits to animate animal faces seamlessly. The model is trained on a hybrid dataset of private Chinese and public English data, enabling multilingual support. Experimental results validate the effectiveness of our approach. Future work will focus on improving real-time performance and refining expression control, further expanding the framework’s applications in portrait animation. The code is available at: \href{https://github.com/jdh-algo/JoyVASA}{\textcolor{pink}{https://jdh-algo.github.io/JoyVASA}}.

\end{abstract}


\section{Introduction}

In recent years, the domain of audio-driven portrait animation has achieved remarkable progress, largely driven by the emergence of diffusion-based generative models \cite{ye2023geneface++,cui2024hallo2,wei2024aniportrait, jiang2024loopy,corona2024vlogger,tian2024emo,zhang2024emotalker}. These innovative approaches have significantly improved both the quality of generated videos and the accuracy of lip synchronization, contributing to more lifelike and engaging animated characters. This technology finds application across diverse fields such as digital avatars \cite{bertoa2020digital}, virtual assistants \cite{song2022talking, curtis2021improving}, and entertainment \cite{zhao2023chatanything, wan2024building}, where realistic animation plays a pivotal role in enhancing user engagement. 

Nevertheless, the growing complexity of these models introduces challenges, leading to inefficiencies in the training and inference phases. This added complexity also restricts the length of videos that can be generated in a single batch, which may negatively impact the continuity of motion between frames. For instance, a lot of existing methods predict approximately 10 frames of images, which corresponds to about 0.4 seconds of video when considering a standard frame rate of 25 frames per second \cite{guan2023stylesync, chen2024echomimic,tian2024emo, xu2024hallo, shi2024joyhallo, cui2024hallo2}. This limitation poses challenges for models in capturing motion information from preceding frames, potentially leading to unstable outcomes, such as sudden temporal jitter and significant deformation of the character's appearance. To address these issues, some studies have proposed simply increasing the number of predicted frames as well as the number of preceding frames \cite{jiang2024loopy}. However, this approach introduces a substantial computational burden, complicating the task. Besides, numerous existing methods require the provision of motion guidance or reference actions, such as action sequence images or facial key points. However, the reliance on such additional reference information reduces the flexibility of the model, as the reference actions typically need to be extracted from supplementary videos.

To alleviate these limitations, we introduce a diffusion-based framework for facial dynamics and head motion generation in audio-driven portrait animation. Our proposed approach harnesses the power of disentangled facial representation to generate compact dynamic sequences directly from audio inputs, without any other conditions. This enables the production of longer animated videos, surpassing the capabilities of current end-to-end diffusion models. Specifically, our framework utilizes a decoupled facial representation model, Liveportrait \cite{guo2024liveportrait}, which distinctly separates dynamic facial expressions from static 3D facial representations. This separation allows for the flexible amalgamation of static representations with dynamically generated sequences, yielding more accurate and adaptable animations. Furthermore, we develop a diffusion transformer model that synthesizes motion sequences, including dynamic facial expressions and head motions. This model operates independently of character identity, drawing solely on audio cues. This feature enhances the versatility of our method, allowing it to be applied to a broad range of character types, including both human and animal figures. A renderer, trained within the decoupled representation framework, then integrates static 3D facial representations with the generated motion sequences to create high-quality animated outputs. Our framework is trained on a hybrid dataset that combines our private Chinese dataset with two publicly available English datasets, ensuring better multilingual support. The results of our experiments validate the effectiveness of our approach. Subsequent advancements will focus on enhancing real-time processing capabilities and refining the control over character expressions, thus expanding the applicability of our framework within the field of portrait animation.

\begin{figure}[!t]
    \centering
    \includegraphics[width=0.95\linewidth]{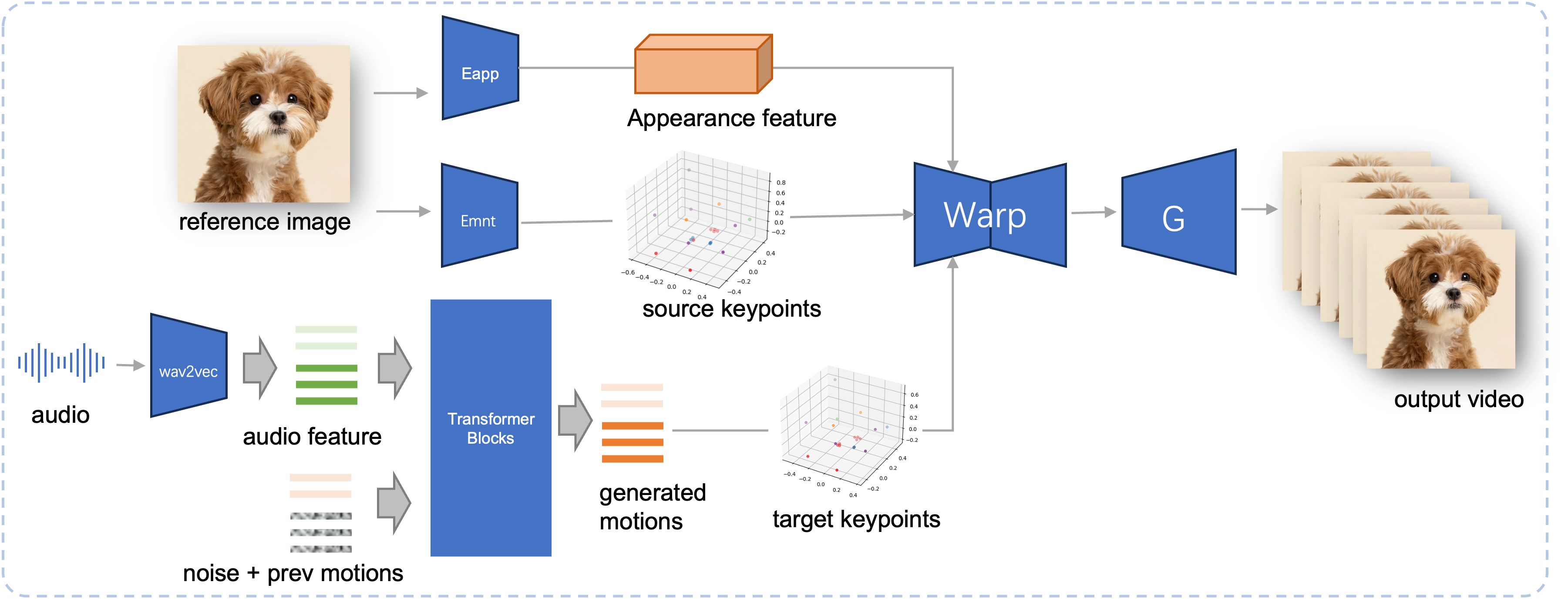}
    \caption{Inference Pipeline of the proposed JoyVASA. }
    \label{fig:pipeline-inference}
\end{figure}

\section{Related Works}\label{sec:related_works}

\textbf{Audio-Driven portrait Animation:} Significant advancements have been made in the field of audio-driven portrait animation in the past few years. Existing methods can be broadly categorized into two main approaches. The first category involves end-to-end audio-to-video generation without any intermediate representations \cite{prajwal2020lip, zhang2023dinet, guan2023stylesync, tian2024emo, chen2024echomimic, xu2024hallo, shi2024joyhallo, park2022synctalkface}. The second category introduces intermediate representations, such as motion sequences or keypoints, as a transitional step before generating the final video \cite{zhang2023sadtalker,ma2023dreamtalk, sun2023vividtalk,liu2023moda}. These methodologies are referred to herein as two-stage methods.

End-to-end approaches typically generate output videos directly from the input audio. Early GAN-based models, such as Wav2Lip \cite{prajwal2020lip}, Dinet \cite{zhang2023dinet}, and StyleSync \cite{guan2023stylesync}, have focused on achieving high-quality lip synchronization. Generally, there lacks a strong correlation between the audio and the non-verbal aspects of facial expression and head movements. Consequently, the weak correlation between audio and motion limits their ability to generate dynamic and expressive cues. Recent advancements have introduced diffusion-based methods, such as EMO \cite{tian2024emo}, Hallo \cite{xu2024hallo}, EchoMimic \cite{chen2024echomimic}, which incorporate motion sequences, speed embeddings, or facial landmarks as constraints to enhance the expressiveness of the generated videos. These additional conditions enable not only lip synchronization but also more vivid and realistic motion generation. However, such methods come with increased computational costs and usually require extra conditional inputs, like reference motion frames, which reduce their flexibility. Moreover, both GAN-based and diffusion-based approaches face challenges in disentangling specific facial expressions or controlling attributes like eye gaze direction and emotional states, as facial representations are encoded as a unified embedding without explicit disentanglement.

Two-stage approaches address some of the aforementioned limitations by introducing facial representation as an intermediate step. Typically, these methods first employ an audio-to-motion model to generate intermediate results, which are then used by a motion-to-video model to render the final video based on the generated motions and the input portrait. Earlier models map the input audio into landmarks \cite{chen2019hierarchical} or 3DMMs\cite{thies2020neural,yi2020audiodriven}, and then generate the final animated videos. Recent GAN-based models such as SadTalker \cite{zhang2023sadtalker} generates motion sequences that are rendered using frameworks like FaceVid2Vid \cite{wang2021one}. The latest innovations leverage diffusion models in a two-stage process: for instance, DreamTalk \cite{ma2023dreamtalk} generates motion sequences utilized by PIRender \cite{ren2021pirenderer} for rendering, while VASA-1 \cite{xu2024vasa} generates motions that are rendered by Megaportrait \cite{drobyshev2022megaportraits}. The introduction of facial representations as an intermediate step, together with the diffusion models, allows two-stage approaches to offer more flexible control over facial expressions and head movements, leading to more realistic and coherent facial expressions and motions.

\textbf{Disentangled Facial Representation}

In recent years, facial representation learning has been explored extensively, especially in disentangling face representation into various attributes. Traditional non-learnable parametric models, such as the widely used 3D Morphable Model (3DMM) \cite{blanz2023morphable,amberg2008expression}, represent faces as a linear combination of identity and expression bases. More recently, non-linear parametric models like FLAME \cite{li2017learning} extend this approach by incorporating articulated attributes such as the jaw, neck, and eyes. These linear or non-linear models are limited in their expressiveness and disentanglement capabilities, often struggling to represent complex and subtle facial motions \cite{jiang2019disentangled}. With the rapid advancements in deep learning, learning-based approaches have been introduced to map 3D face shapes into nonlinear parameter spaces, significantly improving the representational power of these models \cite{drobyshev2024emoportraits,wang2021one,guo2024liveportrait, drobyshev2022megaportraits}. For example, approaches such as the First Order Motion Model (FOMM) \cite{siarohin2019first}, LivePortrait \cite{guo2024liveportrait}, and EmoPortrait \cite{drobyshev2024emoportraits} leverage deep networks to encode detailed geometric features and facial movements. These methods are able to represent intricate details and complex facial dynamics using fewer parameters, offering more flexibility in reconstructing and animating facial expressions.

\section{Method}
\label{sec:method}

\subsection{Overview}
The training process is divided into two main stages: the first stage is disentangling facial representation and the second stage generating audio-conditioned motion. In the first stage, the facial representation is decomposed into two distinct components: 3D appearance features and motion features. The 3D appearance features capture the static characteristics of the face, reflecting the unique identity and visual traits of an individual. In contrast, the motion features encode dynamic elements, including facial expressions, scaling, rotation, and translation, with a primary focus on head movements. This decomposition allows for a clear separation of static and dynamic aspects of facial representation, which enables more flexible portrait animation. In the second stage, a diffusion model is trained to generate motion features conditioned on audio input. This model produces motion features that are independent of the individual’s identity, allowing for the synthesis of audio-driven talking faces. The final output is rendered using the generated motion features and the 3D appearance features obtained from the first stage.

\subsection{Disentangling Facial Representation}

Facial representation disentangling aims to separate different aspects of facial data into distinct components, such as appearance features (e.g., facial structure and texture) and motion attributes (e.g., expressions, head movements). Isolating these elements enables precise control over expressions and movements driven by audio, as appearance remains static while motion can adapt dynamically to audio cues. This separation also reduces the risk of undesired visual distortions, enhancing realism. Additionally, disentangled models allow for reusability, where the same appearance features can pair with various audio inputs to produce diverse animations, boosting efficiency and flexibility.

We use the decoupled facial representation by introducing the existing facial reenactment framework Liveportrait \cite{ guo2024liveportrait}. Formally, given a source image $s$ and a driving video sequence $d_1, d_2, ..., d_N$, this framework first extracts the appearance features and motion information from the source image and each frame of the driving video, then uses the motion information extracted from the driving video to animate the source image. The facial representation is divided into a 3D facial embedding $f_{face}$ and motion features. The 3D facial embedding is extracted by a learned appearance encoder $E_{app}$, and the motion features are extracted by a learned motion encoder $E_mnt$. The motion features is characterized by keypoint displacements ($\delta$) in the canonical space, a rotation matrix ($R$), a translation vector ($t$), and a scaling factor ($s$). Then, a transformation is formalized by the following equation:

\begin{equation}
\label{equ:facial_representation}
\begin{aligned}
    x_{s} &= T(x_{c}, s_{s}, R_{s}, t_{s}, \delta_{s}) = s_{s} \cdot (x_{c}R_{s} + \delta_{s}) + t_{s}, \\
    x_{d} &= T(x_{c}, s_{d}, R_{d}, t_{d}, \delta_{d}) = s_{d} \cdot (x_{c}R_{d} + \delta_{d}) + t_{d}
\end{aligned}
\end{equation}

where $x_{c}$ is the 3D keypoints extracted using a cononical 3D keypoint detection network. $x_{s}$ is the source 3D keypoints deformed using the cononical source 3D keypoints $x_{c}$ and the source motion information ($R_{s}$, $t_s$, $\delta_{s}$, $s_{s}$) extract from the source image, and $x_{d}$ is the drived 3D keypoints transformated using the cononical source 3D keypoints $x_{c}$ and the driving motion information ($R_{d}$, $t_d$, $\delta_{d}$, $s_{d}$) extract from the driving video frame. Readers are referred to \cite{guo2024liveportrait} for more details of this framework.

\subsection{Audio-Driven Motion Sequence Generation with Diffusion}

\begin{figure}
    \centering
    \includegraphics[width=0.95\linewidth]{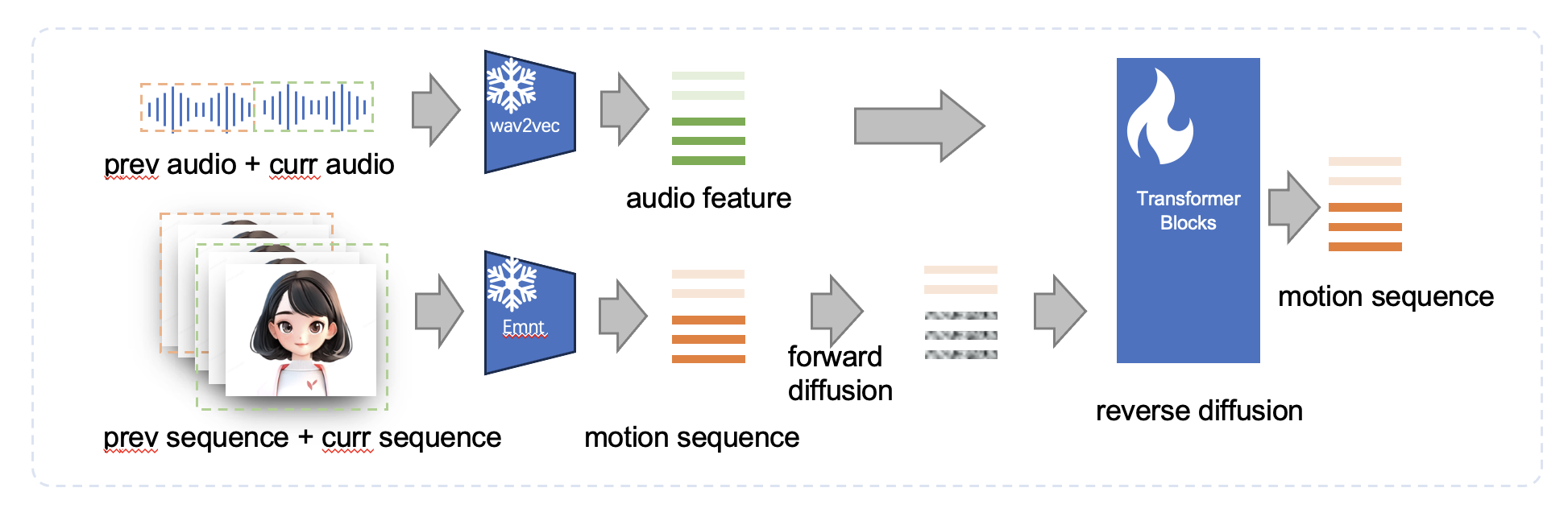}
    \caption{Training process of the audio-driven motion sequence generation. The audio feature and real motion sequences are first extracted with the frozen wav2vec2 \cite{baevski2020wav2vec} and the frozen motion encoder in Liveportrait \cite{guo2024liveportrait}. Then a diffusion transformer model is trained to sample the clean motion sequence from noise.}
    \label{fig:pipeline-train}
\end{figure}

The primary objective of the audio-driven motion sequence generation using the diffusion module is to create a comprehensive sequence of motion information. It is essential to note that, in contrast to some existing methods \cite{zhang2023sadtalker}, the term "motion information" in this context encompasses not only head movements but also all relevant facial dynamics in the first-stage face representation. Thus, our diffusion model aims to generate a holistic facial dynamics representation and the corresponding head movement, akin to that proposed in VASA-1 \cite{xu2024vasa}. Moreover, we exclude any information derived from images during the motion sequence generation process. This strategy is intended to promote the model's ability to learn a more generalized mapping between audio input and facial motion information. By not relying on image data, we also effectively reduce the training and inference costs associated with the model, leading to a more efficient implementation. 

\subsubsection{Audio Conditioned Diffusion Model}

Similar to VASA-1, once the motion encoder \( E_{mnt} \) is trained during the disentangling facial representation stage, we can extract motion sequences from real-life videos to train the diffusion transformer. Specifically, we leverage a frozen wav2vec2 encoder \cite{baevski2020wav2vec} for audio feature extraction, while the frozen motion encoder from LivePortrait \cite{guo2024liveportrait} is utilized to extract motion sequences. The motion sequences \( X_i = \{[R_i, t_i, \delta_i, s_i]\}, i = 1, 2, \dots, N \) are used as input to the diffusion transformer, with the corresponding audio features \( A^i = \{f_{audio}^i\} \) serving as conditioning data. Here, \( N \) represents the number of frames in the real-life video. Figure \ref{fig:pipeline-train} provides an illustration of the training process for generating audio-driven facial dynamic sequences.

In the diffusion model, we aim to sample predicted motions \( \hat{X}^0 \) from noisy observations \( X^t \), conditioned on the corresponding audio features \( A \), where \( t \in T \) denotes the discrete time steps in the diffusion process. To improve temporal consistency in motion generation, we adopt a strategy that includes both the previous and current windows of features in the model’s training. Specifically, at each diffusion step \( t \), the network takes as input both the past and present speech features \( [A_{-W_{pre}} : A_{W_{cur}}] \), the previous motion parameters \( X_{-W_{pre}}^0 \), and the current noisy motion parameters \( X_{W_{cur}}^t \), which are sampled from \( q(X^t_{0:W_{cur}} | X^0_{0:W_{cur}}) \). The denoising network then generates the clean motion samples, as described by the following equation:

\begin{equation}
    \label{equ:reverse_precess}
    \hat{X}^0_{-W_{pre}:W_{cur}} = D(X^t_{0:W_{pre}}, X^0_{-W_{pre}:0}, A_{-W_{pre}:W_{cur}}, t)
\end{equation}

For the initial window, the speech features and motion parameters are substituted with learnable start features \( A_{start} \) and \( X_{start} \), respectively. It is important to note that the conditioning used in our approach can be expressed as \( C = [X_{-W_{pre}:0}, A_{-W_{pre}:W_{cur}}] \), which is more compact compared to VASA-1. In this formulation, the gaze direction and head-to-camera distance, encoded in VASA-1 as separate features, are instead represented by the motion parameters \( \delta \) and \( s \), respectively. Generally, the length of the audio may not fully occupy the entire window during training. To improve the model's robustness to varying speech lengths, we introduce random truncation of the samples during training, ensuring the model can handle audio sequences of different durations effectively.

Classifier-free guidance enhances generation quality by directly conditioning the model on the desired output, eliminating the need for an additional classifier. This approach enables more flexible control over the generated samples while preserving high fidelity. In this paper, we apply classifier-free guidance during the motion sequence generation process, which can be expressed as follows:

\begin{equation}
\begin{aligned}
    \hat{X}^{0} = & D (X^t_{0:W_{pre}}, X^0_{-W_{pre}:0}, \emptyset, t) + \\
    & \lambda_c ( D(X^t_{0:W_{pre}}, X^0_{-W_{pre}:0}, A_{-W_{pre}} : A_{W_{cur}}, t) - \\
    & D(X^t_{0:W_{pre}}, X^0_{-W_{pre}:0}, \emptyset, t) )
\end{aligned}
\end{equation}

Where $\lambda_c$ is the CFG scale, and the audio condition is randomly set to $\emptyset$ with 0.1 probability.

\subsubsection{Loss Functions}

To impose stronger constraints on the model predictions, we directly sample the clean motion sequence rather than noise in the diffusion model. In this work, we utilize three distinct loss functions: the simple loss, velocity loss \cite{cudeiro2019capture}, and smooth loss \cite{sun2024diffposetalk}. The simple loss \( L_{simple} \) is defined as the \( L_2 \) distance between the real motion sequences \( X^0_{-W_{pre}:W_{cur}} \) and the generated clean motion sequence \( \hat{X}^0_{-W_{pre}:W_{cur}} \). The velocity loss \( L_{vel} \) is introduced to encourage improved temporal consistency, while the smooth loss \( L_{smooth} \) penalizes large accelerations in the predicted motions. Additionally, we incorporate an expression loss \( L_{exp} \) to place greater emphasis on lip movements and facial expressions in the generated sequences. These loss functions are mathematically defined as follows:

\begin{equation}
    L_{sample} = || X^{0}_{-W_{pre}:W_{cur}} - \hat{X}^{0}_{-W_{pre}:W_{cur}} ||^{2}
\end{equation}

\begin{equation}
    L_{vel} = || (X^{0}_{-W_{pre}+1:W_{cur}} - X^{0}_{-W_{pre}:W_{cur}-1}) - (\hat{X}^{0}_{-W_{pre}+1:W_{cur}} - \hat{X}^{0}_{-W_{pre}:W_{cur}-1}) ||^2
\end{equation}

\begin{equation}
    L_{smooth} = || (\hat{X}^{0}_{-W_{pre}+2:W_{cur}} - 2\hat{X}^{0}_{-W_{pre}+1:W_{cur}-1}) + \hat{X}^{0}_{-W_{pre}:W_{cur}-2} ||^2
\end{equation}

\begin{equation}
    L_{exp} = || \delta^{0}_{-W_{pre}:W_{cur}} - \hat{\delta}^{0}_{-W_{pre}:W_{cur}} ||^{2}
\end{equation}

The total loss used in this paper can be formulated as:

\begin{equation}
    L_{total} = L_{simple} + \lambda_{vel} * L_{vel} + \lambda_{smooth} * L_{smooth} + \lambda_{exp} * L_{exp}
\end{equation}

where \( \lambda_{vel} \), \( \lambda_{smooth} \), and \( \lambda_{exp} \) are the weights for the velocity loss, smooth loss, and expression loss, respectively.

\subsection{Inference}
The inference pipeline is illustrated in Figure \ref{fig:pipeline-inference}. Given a reference image, we first extract the corresponding 3D facial appearance feature \( f_{face} \) using the appearance encoder \( E_{app} \), and the learned motion information \( [x_c, R_s, \delta_s, t_s] \) using the motion encoder \( E_{mnt} \). For the input speech, the audio features are initially extracted using the wav2vec2 encoder. The audio-driven motion sequences \( \hat{X}^0 \) are then sampled using the diffusion model trained in the second stage in a sliding window fashion. Using the canonical source keypoints \( x_c \) and the sampled target motion sequences \( \hat{X}^0 \), the target keypoints \( x_d \) are computed as shown in equation \ref{equ:facial_representation}. Finally, the 3D facial appearance feature is warped based on the source and target keypoints and rendered by a generator to produce the final output video.

\section{Experiments}

\subsection{Datasets}

Three datasets were employed to train the motion generation model: the publicly available HDTF dataset \cite{Zhang2021flow_guided}, the CelebV-HQ dataset \cite{zhu2022celebv}, and a proprietary high-resolution speaking dataset collected by our team. To ensure data quality, a video quality assessment model \cite{wu2023qalign} was implemented to exclude low-quality samples across all datasets. Additionally, a lip synchronization model \cite{Chung2016OutOT} was used to filter out videos with low synchronization scores, further refining the dataset. We further remove clips with excessive head motion to avoid distortion in the training process. Ultimately, we obtained a total of 5,578 video clips in the training dataset, with individual clip durations ranging from 8 seconds to several minutes. It is worth noting that an oversampling strategy was applied to the training dataset to balance the sample sizes across the different datasets, ensuring that each contributed a comparable number of samples.
For model evaluation, two distinct test sets were employed. The first, referred to as the celebV-HQ test dataset, is a subset of the celebV-HQ dataset. It comprises 50 randomly selected individuals, each represented by two video clips with durations ranging from 5 to 15 seconds. It is important to note that none of the celebV-HQ test videos appear in the training dataset. The second, termed the Openset dataset, includes 50 images sourced from the internet, featuring real individuals, animated characters, and various handcrafted materials. These images are paired with 50 corresponding audio samples, covering a diverse range of content, such as speeches, emotional dialogues, and singing performances.

\subsection{Implementation Details}
The diffusion transformer is trained using an NVIDIA A800 GPU. The motion generation architecture consists of a six-layer transformer decoder with eight attention heads for the denoising network, featuring a dimensionality of 512. The window length \(W_{cur}\) is set to 100, while \(W_{pre}\) is configured to 25. For optimization, we utilize the Adam optimizer, conducting a total of 20,000 training steps with a batch size of 16 and a learning rate of \(1 \times 10^{-4}\). The \( \lambda_{vel} \), \( \lambda_{smooth} \), and \( \lambda_{exp} \) are set to 5.0, 0.5 and 0.1, respectively.

\subsection{Evaluation Metrics}
We employed several evaluation metrics to assess the performance of the proposed method. The widely used sync-C and sync-D metrics were introduced to evaluate the audio-lip synchronization across different approaches. To assess the quality of the generated videos, both image quality assessment (IQA) \cite{wu2023qalign} and video quality assessment (VQA) metrics \cite{wu2023qalign} were utilized. Additionally, a temporal smoothness metric was employed to measure the continuity of motion in the generated videos. For the HDTF test dataset, which includes corresponding ground truth videos, we also used the Fréchet Video Distance (FVD) \cite{fvd_2019} and Fréchet Inception Distance (FID) metrics to further evaluate the fidelity of the generated outputs.

\section{Results}
\subsection{Visualization results}

\begin{figure}
    \centering
    \includegraphics[width=0.99\linewidth]{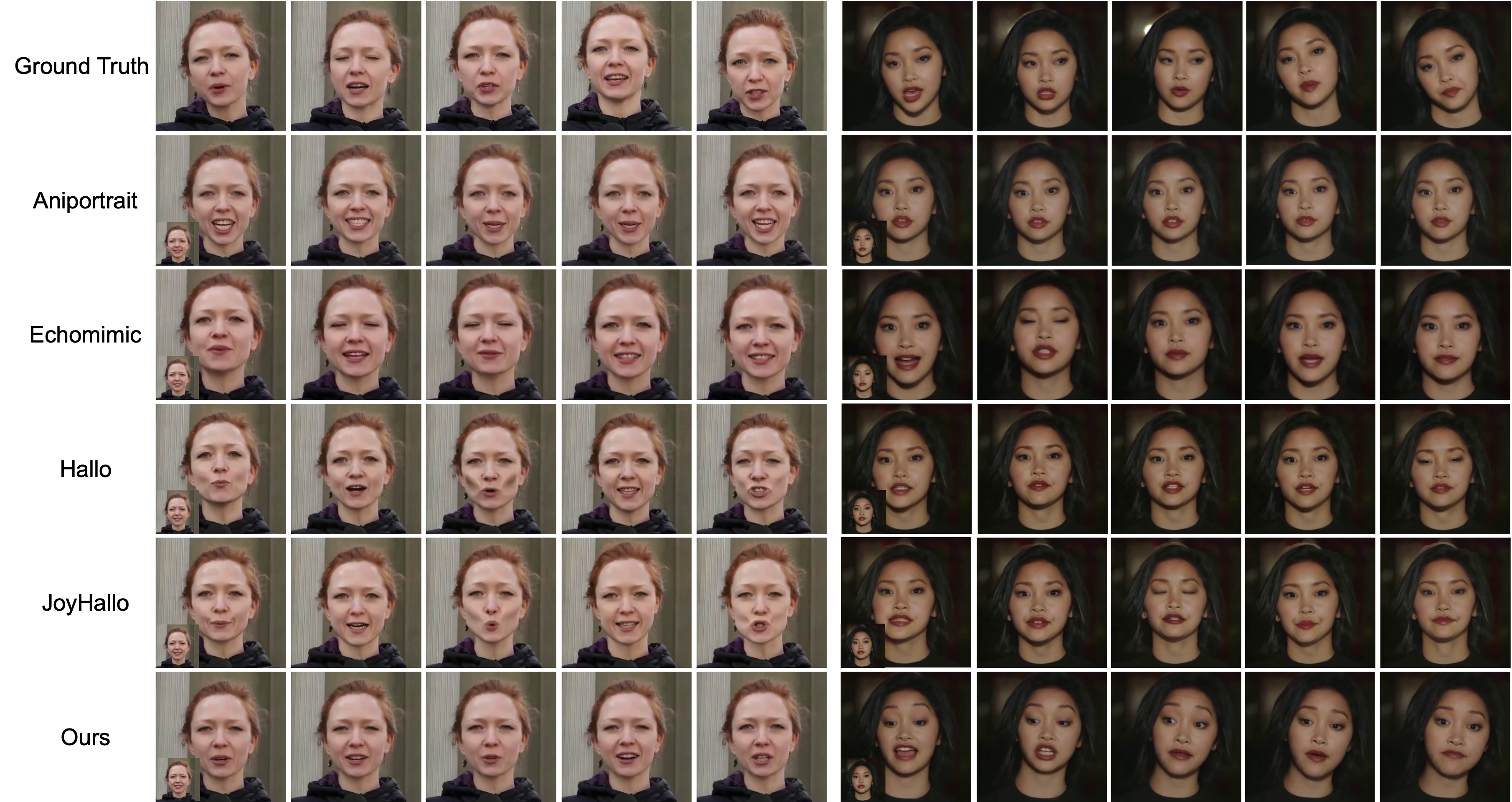}
    \caption{Visualization results of different methods on the celebV-HQ test dataset. }
    \label{fig:result_diff_methods}
\end{figure}

Figure \ref{fig:result_diff_methods} presents a comparison of the visualization results between existing portrait animation methods and our proposed approach, using the celebV-HQ test dataset. For all methods, the first frame and audio track from the ground truth video are utilized as the reference image and driving audio, respectively. The lower-left corner of the first column shows the initial frame used by all methods for reference.Overall, our method demonstrates competitive performance. As shown in the example on the left, although the amplitude of the mouth movement in our method is less pronounced compared to some of the baseline methods (e.g., Aniportrait, Hallo), the mouth shape produced by our approach more closely aligns with the ground truth. The figure on the right illustrates that our method generates more expressive head movements compared to other techniques.

\begin{figure}
    \centering
    \includegraphics[width=0.95\linewidth]{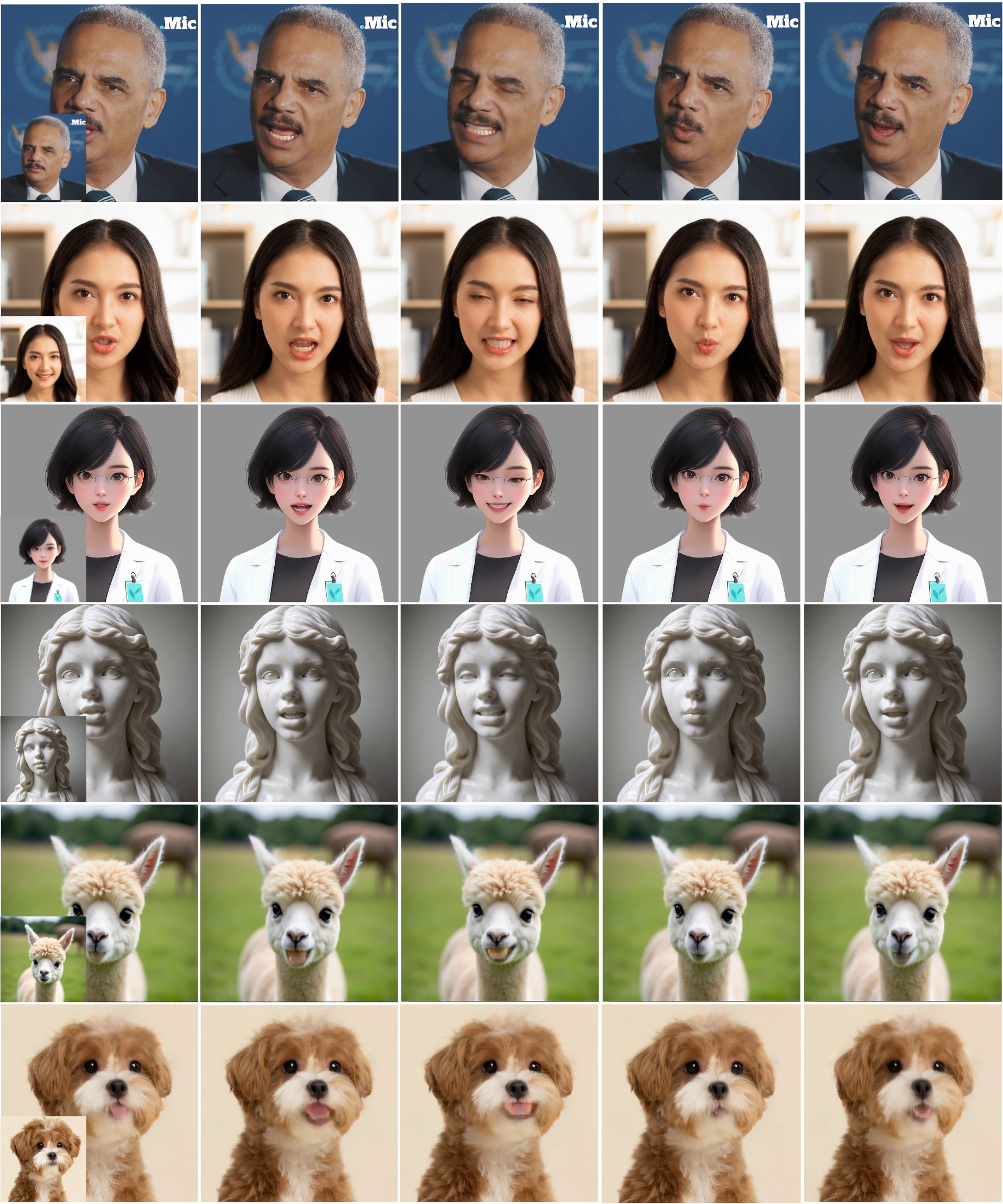}
    \caption{Visualization results of different portraits driven by the same audio input on the Openset dataset. Note that our proposed method is able to drive portraits of humans, animations, artworks, and animals at the same time.}
    \label{fig:result_diff_portraits}
\end{figure}

Figure \ref{fig:result_diff_portraits} shows the results of different portraits driven by the same input audio on the Openset dataset. It can be seen that our results exhibit more diverse facial expressions and a wider range of lip movements. Since the motion sequences we predict are only related to audio, the predicted model can drive any portrait, including people, anime, portraits with artwork, and even animals. Although this may cause the problem that all characters end up having the same or similar expressions, by adjusting the CFG weight, we can easily obtain results with other expressions and movements.

\subsection{Comparison with existing methods}
\begin{table}[!t]
    \caption{Comparison with existing methods on the celebv-HQ test dataset. $\uparrow$ larger is better. $\downarrow$ smaller is better.}
    \centering
    \begin{tabular}{ccccccc}
        \toprule
        Method & IQA(\%) $\uparrow$ & VQA(\%) $\uparrow$ & Sync-C $\uparrow$ & Sync-D $\downarrow$ & FVD-25$\downarrow$ & Smooth(\%) $\uparrow$ \\
        \midrule
        EchoMimic \cite{chen2024echomimic} & 64.46 & 68.85 & 4.60 & 13.90 & 904.07 & 99.44 \\
        Aniportrait \cite{wei2024aniportrait} & \textbf{74.85} & \textbf{78.00} & 1.98 & \textbf{13.28} & 623.00 & \textbf{99.66} \\
        Hallo \cite{xu2024hallo}  & 68.83 & 72.95 & 6.25 & 14.05 & 558.96 & 99.47 \\
        JoyHallo \cite{shi2024joyhallo} & 69.15 & 74.00 & \textbf{6.36} & 14.23 & 755.61 & 99.38 \\
        Ours & 68.97 & 72.42 & 4.85 & 13.53 & \textbf{459.04} & 99.60 \\
        \bottomrule
    \end{tabular}
    \label{tab:hdtf_testset}
\end{table}

Table \ref{tab:hdtf_testset} compares our method with existing techniques on the celebvHQ test dataset across various metrics. In terms of Image Quality Assessment (IQA) and Video Quality Assessment (VQA), our method achieves 68.97\% and 72.42\%, respectively, which is slightly lower than Aniportrait’s 74.85\% and 78.00\%, but still ranks above average. Additionally, in Sync-C (synchronization consistency), our method (4.85) performs similarly to EchoMimic (4.60) but lags behind Aniportrait’s 1.98, indicating a slightly lower level of synchronization in the generated animations. For Sync-D (synchronization delay), our method (13.53) is close to Aniportrait’s 13.28 and performs comparably to other methods in this regard. Notably, our method achieves the lowest FVD-25 score (459.04) among all the methods, indicating superior video generation quality in terms of visual smoothness and naturalness. In the Smoothness metric, our method (99.60\%) outperforms all other methods except Aniportrait (99.66\%). Overall, while our method may not outperform Aniportrait in every metric, it excels in FVD-25 and Smoothness, showcasing its strong capabilities in generating high-quality videos.

\begin{table}[!t]
    \caption{Comparison with existing methods on Openset dataset. $\uparrow$ larger is better. $\downarrow$ smaller is better.}
    \centering
    \begin{tabular}{cccccc}
        \toprule
        Method & IQA $\uparrow$ & VQA $\uparrow$ & Sync-C $\uparrow$ & Sync-D $\downarrow$ & Smooth $\uparrow$ \\
        \midrule
        EchoMimic \cite{chen2024echomimic} & 75.66 & 82.73 & 4.12 & 13.79 & 99.20 \\
        Aniportrait \cite{wei2024aniportrait} & \textbf{81.84} & \textbf{87.62} & 2.67 & 14.09 & \textbf{99.55} \\
        Hallo \cite{xu2024hallo} &  75.37 & 83.85 & 5.39 & 13.91 & 99.33 \\
        JoyHallo \cite{shi2024joyhallo} & 73.87 & 84.09 & \textbf{5.91} & 14.11  & 99.32 \\
        Ours & 71.45 & 77.78 & 5.72 & 14.01 & 99.48 \\
        \bottomrule
    \end{tabular}
    \label{tab:openset_dataset}
\end{table}

Table \ref{tab:openset_dataset} compares our method with existing approaches on the Openset dataset. While our method shows competitive performance, it slightly lags behind Aniportrait in IQA (71.45 vs. 81.84) and VQA (77.78 vs. 87.62). For Sync-C, our method achieves a score of 5.72, higher than Aniportrait’s 2.67, indicating a minor tradeoff in synchronization accuracy. However, our method performs comparably in Sync-D (14.01) and Smoothness (99.48), closely matching the best results. Overall, our method balances synchronization, smoothness, and quality, offering promising performance with some room for improvement in visual quality and synchronization accuracy.




\section{Conclusion and Future Work}
In conclusion, this paper introduces a novel diffusion-based framework for audio-driven portrait animation that effectively addresses challenges related to video quality and lip synchronization. By employing a disentangled facial representation that separates dynamic expressions from static features, our approach enables the generation of longer and more coherent animated videos. The integration of a diffusion transformer allows for character-independent facial representation synthesis based solely on audio input, broadening the method's applicability to both human and animal faces with the use of Liveportrait. Our experiments validate the framework's effectiveness, demonstrating high-quality output and robust multilingual support through a hybrid dataset. 

One limitation of this work lies in the upper bound of the generated content, which is constrained by the performance of the first-stage model, including the appearance encoder, motion encoder, and decoder trained in this phase. Specifically, the Liveportrait model used in the first stage exhibits suboptimal performance when handling large pose variations, which in turn affects the quality of the generated outputs. Furthermore, Liveportrait incorporates a retargeting module designed for cross-identity reference images, which is not applicable in the context of pure audio-driven portrait animation. Consequently, this limitation also impacts the quality of the generated results. In future work, we aim to explore more robust face representation techniques, such as EMOPortrait, to address these issues. Moreover, we will focus on enhancing the real-time processing capabilities of the model and refining the control over facial expressions. These improvements will expand the potential applications of this approach in portrait animation, making it more versatile and suitable for a wider range of use cases.

\bibliographystyle{unsrt}
\bibliography{main.bib}  

@article{ye2023geneface++,
  title={Geneface++: Generalized and stable real-time audio-driven 3d talking face generation},
  author={Ye, Zhenhui and He, Jinzheng and Jiang, Ziyue and Huang, Rongjie and Huang, Jiawei and Liu, Jinglin and Ren, Yi and Yin, Xiang and Ma, Zejun and Zhao, Zhou},
  journal={arXiv preprint arXiv:2305.00787},
  year={2023}
}

@inproceedings{zhang2024emotalker,
  title={Emotalker: Emotionally editable talking face generation via diffusion model},
  author={Zhang, Bingyuan and Zhang, Xulong and Cheng, Ning and Yu, Jun and Xiao, Jing and Wang, Jianzong},
  booktitle={ICASSP 2024-2024 IEEE International Conference on Acoustics, Speech and Signal Processing (ICASSP)},
  pages={8276--8280},
  year={2024},
  organization={IEEE}
}

@inproceedings{zhang2023sadtalker,
  title={Sadtalker: Learning realistic 3d motion coefficients for stylized audio-driven single image talking face animation},
  author={Zhang, Wenxuan and Cun, Xiaodong and Wang, Xuan and Zhang, Yong and Shen, Xi and Guo, Yu and Shan, Ying and Wang, Fei},
  booktitle={CVPR},
  pages={8652--8661},
  year={2023}
}

@article{guo2024liveportrait,
  title={Liveportrait: Efficient portrait animation with stitching and retargeting control},
  author={Guo, Jianzhu and Zhang, Dingyun and Liu, Xiaoqiang and Zhong, Zhizhou and Zhang, Yuan and Wan, Pengfei and Zhang, Di},
  journal={arXiv preprint arXiv:2407.03168},
  year={2024}
}

@inproceedings{drobyshev2022megaportraits,
  title={Megaportraits: One-shot megapixel neural head avatars},
  author={Drobyshev, Nikita and Chelishev, Jenya and Khakhulin, Taras and Ivakhnenko, Aleksei and Lempitsky, Victor and Zakharov, Egor},
  booktitle={Proceedings of the 30th ACM International Conference on Multimedia},
  pages={2663--2671},
  year={2022}
}

@inproceedings{drobyshev2024emoportraits,
  title={EMOPortraits: Emotion-enhanced Multimodal One-shot Head Avatars},
  author={Drobyshev, Nikita and Casademunt, Antoni Bigata and Vougioukas, Konstantinos and Landgraf, Zoe and Petridis, Stavros and Pantic, Maja},
  booktitle={CVPR},
  pages={8498--8507},
  year={2024}
}

@inproceedings{Zhang2021flow_guided,  
 title={Flow-guided One-shot Talking Face Generation with a High-resolution Audio-visual Dataset}, 
 url={http://dx.doi.org/10.1109/cvpr46437.2021.00366}, 
 DOI={10.1109/cvpr46437.2021.00366}, 
booktitle={CVPR},
 author={Zhang, Zhimeng and Li, Lincheng and Ding, Yu and Fan, Changjie}, 
 year={2021}, 
 month={Jun}, 
 language={en-US} 
 }

@inproceedings{zhu2022celebv,
  title={CelebV-HQ: A large-scale video facial attributes dataset},
  author={Zhu, Hao and Wu, Wayne and Zhu, Wentao and Jiang, Liming and Tang, Siwei and Zhang, Li and Liu, Ziwei and Loy, Chen Change},
  booktitle={ECCV},
  pages={650--667},
  year={2022},
  organization={Springer}
}

@article{wu2023qalign,
  title={Q-align: Teaching lmms for visual scoring via discrete text-defined levels},
  author={Wu, Haoning and Zhang, Zicheng and Zhang, Weixia and Chen, Chaofeng and Liao, Liang and Li, Chunyi and Gao, Yixuan and Wang, Annan and Zhang, Erli and Sun, Wenxiu and others},
  journal={arXiv preprint arXiv:2312.17090},
  year={2023}
}

@inproceedings{Chung2016OutOT,
  title={Out of Time: Automated Lip Sync in the Wild},
  author={Joon Son Chung and Andrew Zisserman},
  booktitle={ACCV Workshops},
  year={2016},
  url={https://api.semanticscholar.org/CorpusID:26294509}
}

@inproceedings{zhang2023dinet,
  title={Dinet: Deformation inpainting network for realistic face visually dubbing on high resolution video},
  author={Zhang, Zhimeng and Hu, Zhipeng and Deng, Wenjin and Fan, Changjie and Lv, Tangjie and Ding, Yu},
  booktitle={Proceedings of the AAAI Conference on Artificial Intelligence},
  volume={37},
  number={3},
  pages={3543--3551},
  year={2023}
}

@inproceedings{prajwal2020lip,
  title={A lip sync expert is all you need for speech to lip generation in the wild},
  author={Prajwal, KR and Mukhopadhyay, Rudrabha and Namboodiri, Vinay P and Jawahar, CV},
  booktitle={Proceedings of the 28th ACM international conference on multimedia},
  pages={484--492},
  year={2020}
}

@misc{guan2023stylesync,
      title={StyleSync: High-Fidelity Generalized and Personalized Lip Sync in Style-based Generator}, 
      author={Jiazhi Guan and Zhanwang Zhang and Hang Zhou and Tianshu Hu and Kaisiyuan Wang and Dongliang He and Haocheng Feng and Jingtuo Liu and Errui Ding and Ziwei Liu and Jingdong Wang},
      year={2023},
      eprint={2305.05445},
      archivePrefix={arXiv},
      primaryClass={cs.CV},
      url={https://arxiv.org/abs/2305.05445}, 
}

@misc{tian2024emo,
      title={EMO: Emote Portrait Alive - Generating Expressive Portrait Videos with Audio2Video Diffusion Model under Weak Conditions}, 
      author={Linrui Tian and Qi Wang and Bang Zhang and Liefeng Bo},
      year={2024},
      eprint={2402.17485},
      archivePrefix={arXiv},
      primaryClass={cs.CV}
}

@misc{chen2024echomimic,
  title={EchoMimic: Lifelike Audio-Driven Portrait Animations through Editable Landmark Conditioning},
  author={Zhiyuan Chen and Jiajiong Cao and Zhiquan Chen and Yuming Li and Chenguang Ma},
  year={2024},
  archivePrefix={arXiv},
  primaryClass={cs.CV}
}

@misc{xu2024hallo,
  title={Hallo: Hierarchical Audio-Driven Visual Synthesis for Portrait Image Animation},
  author={Mingwang Xu and Hui Li and Qingkun Su and Hanlin Shang and Liwei Zhang and Ce Liu and Jingdong Wang and Yao Yao and Siyu zhu},
  year={2024},
  eprint={2406.08801},
  archivePrefix={arXiv},
  primaryClass={cs.CV}
}

@inproceedings{wang2021one,
  title={One-shot free-view neural talking-head synthesis for video conferencing},
  author={Wang, Ting-Chun and Mallya, Arun and Liu, Ming-Yu},
  booktitle={Proceedings of the IEEE/CVF conference on computer vision and pattern recognition},
  pages={10039--10049},
  year={2021}
}

@article{ma2023dreamtalk,
  title={Dreamtalk: When expressive talking head generation meets diffusion probabilistic models},
  author={Ma, Yifeng and Zhang, Shiwei and Wang, Jiayu and Wang, Xiang and Zhang, Yingya and Deng, Zhidong},
  journal={arXiv preprint arXiv:2312.09767},
  year={2023}
}

@inproceedings{ren2021pirenderer,
  title={Pirenderer: Controllable portrait image generation via semantic neural rendering},
  author={Ren, Yurui and Li, Ge and Chen, Yuanqi and Li, Thomas H and Liu, Shan},
  booktitle={Proceedings of the IEEE/CVF international conference on computer vision},
  pages={13759--13768},
  year={2021}
}

@article{xu2024vasa,
  title={Vasa-1: Lifelike audio-driven talking faces generated in real time},
  author={Xu, Sicheng and Chen, Guojun and Guo, Yu-Xiao and Yang, Jiaolong and Li, Chong and Zang, Zhenyu and Zhang, Yizhong and Tong, Xin and Guo, Baining},
  journal={arXiv preprint arXiv:2404.10667},
  year={2024}
}

@inproceedings{chen2019hierarchical,
  title={Hierarchical Cross-Modal Talking Face Generation with Dynamic Pixel-Wise Loss},
  author={Chen, Lele and Maddox, Ross K and Duan, Zhiyao and Xu, Chenliang},
  booktitle={CVPR},
  pages={7832--7841},
  year={2019}
}

@inproceedings{thies2020neural,
  title={Neural voice puppetry: Audio-driven facial reenactment},
  author={Thies, Justus and Elgharib, Mohamed and Tewari, Ayush and Theobalt, Christian and Nie{\ss}ner, Matthias},
  booktitle={Computer Vision--ECCV 2020},
  pages={716--731},
  year={2020},
  organization={Springer}
}

@misc{yi2020audiodriven,
      title={Audio-driven Talking Face Video Generation with Learning-based Personalized Head Pose}, 
      author={Ran Yi and Zipeng Ye and Juyong Zhang and Hujun Bao and Yong-Jin Liu},
      year={2020},
      eprint={2002.10137},
      archivePrefix={arXiv},
      primaryClass={cs.CV},
      url={https://arxiv.org/abs/2002.10137}, 
}

@incollection{blanz2023morphable,
  title={A morphable model for the synthesis of 3D faces},
  author={Blanz, Volker and Vetter, Thomas},
  booktitle={Seminal Graphics Papers: Pushing the Boundaries, Volume 2},
  pages={157--164},
  year={2023}
}

@inproceedings{amberg2008expression,
  title={Expression invariant 3D face recognition with a morphable model},
  author={Amberg, Brian and Knothe, Reinhard and Vetter, Thomas},
  booktitle={2008 IEEE International Conference on Automatic Face \& Gesture Recognition},
  pages={1--6},
  year={2008},
  organization={IEEE}
}

@article{li2017learning,
  title={Learning a model of facial shape and expression from 4D scans.},
  author={Li, Tianye and Bolkart, Timo and Black, Michael J and Li, Hao and Romero, Javier},
  journal={ACM Trans. Graph.},
  volume={36},
  number={6},
  pages={194--1},
  year={2017}
}

@misc{jiang2019disentangled,
      title={Disentangled Representation Learning for 3D Face Shape}, 
      author={Zi-Hang Jiang and Qianyi Wu and Keyu Chen and Juyong Zhang},
      year={2019},
      eprint={1902.09887},
      archivePrefix={arXiv},
      primaryClass={cs.CV},
      url={https://arxiv.org/abs/1902.09887}, 
}

@article{siarohin2019first,
  title={First order motion model for image animation},
  author={Siarohin, Aliaksandr and Lathuili{\`e}re, St{\'e}phane and Tulyakov, Sergey and Ricci, Elisa and Sebe, Nicu},
  journal={Advances in neural information processing systems},
  volume={32},
  year={2019}
}

@article{shi2024joyhallo,
  title={JoyHallo: Digital human model for Mandarin},
  author={Shi, Sheng and Cao, Xuyang and Zhao, Jun and Wang, Guoxin},
  journal={arXiv preprint arXiv:2409.13268},
  year={2024}
}

@article{wei2024aniportrait,
  title={Aniportrait: Audio-driven synthesis of photorealistic portrait animation},
  author={Wei, Huawei and Yang, Zejun and Wang, Zhisheng},
  journal={arXiv preprint arXiv:2403.17694},
  year={2024}
}

@article{baevski2020wav2vec,
  title={wav2vec 2.0: A framework for self-supervised learning of speech representations},
  author={Baevski, Alexei and Zhou, Yuhao and Mohamed, Abdelrahman and Auli, Michael},
  journal={Advances in neural information processing systems},
  volume={33},
  pages={12449--12460},
  year={2020}
}

@inproceedings{fvd_2019,
  author       = {Thomas Unterthiner and
                  Sjoerd van Steenkiste and
                  Karol Kurach and
                  Rapha{\"{e}}l Marinier and
                  Marcin Michalski and
                  Sylvain Gelly},
  title        = {{FVD:} {A} new Metric for Video Generation},
  booktitle    = {Deep Generative Models for Highly Structured Data, {ICLR} 2019 Workshop,
                  New Orleans, Louisiana, United States, May 6, 2019},
  publisher    = {OpenReview.net},
  year         = {2019},
  url          = {https://openreview.net/forum?id=rylgEULtdN},
  bibsource    = {dblp computer science bibliography, https://dblp.org}
}

@article{cui2024hallo2,
  title={Hallo2: Long-Duration and High-Resolution Audio-Driven Portrait Image Animation},
  author={Cui, Jiahao and Li, Hui and Yao, Yao and Zhu, Hao and Shang, Hanlin and Cheng, Kaihui and Zhou, Hang and Zhu, Siyu and Wang, Jingdong},
  journal={arXiv preprint arXiv:2410.07718},
  year={2024}
}

@article{jiang2024loopy,
  title={Loopy: Taming audio-driven portrait avatar with long-term motion dependency},
  author={Jiang, Jianwen and Liang, Chao and Yang, Jiaqi and Lin, Gaojie and Zhong, Tianyun and Zheng, Yanbo},
  journal={arXiv preprint arXiv:2409.02634},
  year={2024}
}

@article{corona2024vlogger,
  title={VLOGGER: Multimodal diffusion for embodied avatar synthesis},
  author={Corona, Enric and Zanfir, Andrei and Bazavan, Eduard Gabriel and Kolotouros, Nikos and Alldieck, Thiemo and Sminchisescu, Cristian},
  journal={arXiv preprint arXiv:2403.08764},
  year={2024}
}

@article{bertoa2020digital,
  title={Digital avatars: Promoting independent living for older adults},
  author={Bertoa, Manuel F and Moreno, Nathalie and Perez-Vereda, Alejandro and Bandera, David and {\'A}lvarez-Palomo, Jos{\'e} M and Canal, Carlos},
  journal={Wireless Communications and Mobile Computing},
  volume={2020},
  number={1},
  pages={8891002},
  year={2020},
  publisher={Wiley Online Library}
}

@inproceedings{song2022talking,
  title={Talking face generation with multilingual tts},
  author={Song, Hyoung-Kyu and Woo, Sang Hoon and Lee, Junhyeok and Yang, Seungmin and Cho, Hyunjae and Lee, Youseong and Choi, Dongho and Kim, Kang-wook},
  booktitle={Proceedings of the IEEE/CVF Conference on Computer Vision and Pattern Recognition},
  pages={21425--21430},
  year={2022}
}

@article{zhao2023chatanything,
  title={ChatAnything: Facetime Chat with LLM-Enhanced Personas},
  author={Zhao, Yilin and Yuan, Xinbin and Gao, Shanghua and Lin, Zhijie and Hou, Qibin and Feng, Jiashi and Zhou, Daquan},
  journal={arXiv preprint arXiv:2311.06772},
  year={2023}
}

@inproceedings{wan2024building,
  title={Building LLM-based AI Agents in Social Virtual Reality},
  author={Wan, Hongyu and Zhang, Jinda and Suria, Abdulaziz Arif and Yao, Bingsheng and Wang, Dakuo and Coady, Yvonne and Prpa, Mirjana},
  booktitle={Extended Abstracts of the CHI Conference on Human Factors in Computing Systems},
  pages={1--7},
  year={2024}
}

@article{curtis2021improving,
  title={Improving user experience of virtual health assistants: scoping review},
  author={Curtis, Rachel G and Bartel, Bethany and Ferguson, Ty and Blake, Henry T and Northcott, Celine and Virgara, Rosa and Maher, Carol A},
  journal={Journal of medical Internet research},
  volume={23},
  number={12},
  pages={e31737},
  year={2021},
  publisher={JMIR Publications Toronto, Canada}
}

@article{sun2023vividtalk,
  title={Vividtalk: One-shot audio-driven talking head generation based on 3d hybrid prior},
  author={Sun, Xusen and Zhang, Longhao and Zhu, Hao and Zhang, Peng and Zhang, Bang and Ji, Xinya and Zhou, Kangneng and Gao, Daiheng and Bo, Liefeng and Cao, Xun},
  journal={arXiv preprint arXiv:2312.01841},
  year={2023}
}

@inproceedings{liu2023moda,
  title={Moda: Mapping-once audio-driven portrait animation with dual attentions},
  author={Liu, Yunfei and Lin, Lijian and Yu, Fei and Zhou, Changyin and Li, Yu},
  booktitle={Proceedings of the IEEE/CVF International Conference on Computer Vision},
  pages={23020--23029},
  year={2023}
}

@inproceedings{park2022synctalkface,
  title={SyncTalkFace: Talking Face Generation with Precise Lip-Syncing via Audio-Lip Memory},
  author={Park, Se Jin and Kim, Minsu and Hong, Joanna and Choi, Jeongsoo and Ro, Yong Man},
  booktitle={AAAI},
  pages={2062--2070},
  year={2022},
  organization={Association for the Advancement of Artificial Intelligence}
}

@inproceedings{cudeiro2019capture,
  title={Capture, learning, and synthesis of 3D speaking styles},
  author={Cudeiro, Daniel and Bolkart, Timo and Laidlaw, Cassidy and Ranjan, Anurag and Black, Michael J},
  booktitle={CVPR},
  pages={10101--10111},
  year={2019}
}

@article{sun2024diffposetalk,
  title={Diffposetalk: Speech-driven stylistic 3d facial animation and head pose generation via diffusion models},
  author={Sun, Zhiyao and Lv, Tian and Ye, Sheng and Lin, Matthieu and Sheng, Jenny and Wen, Yu-Hui and Yu, Minjing and Liu, Yong-jin},
  journal={ACM Transactions on Graphics (TOG)},
  volume={43},
  number={4},
  pages={1--9},
  year={2024},
  publisher={ACM New York, NY, USA}
}






\end{document}